%% file: main.tex
\documentclass[journal]{IEEEtran}

\usepackage{amssymb,amsmath}
\usepackage{cite}
\usepackage{subcaption}
\usepackage{graphicx}
\usepackage{psfrag}
\usepackage{url}
\usepackage{hyperref}
\usepackage[utf8]{inputenc}
\usepackage[absolute,overlay]{textpos}
\usepackage{gensymb}
\usepackage{cases}
\usepackage{bm}
\usepackage{graphicx}
\usepackage[font=footnotesize]{caption}
\usepackage[font=footnotesize]{subcaption}
\usepackage{float}
\usepackage{listings}
\usepackage{color}

\usepackage{tabularx}
\usepackage{array}

\allowdisplaybreaks
\usepackage{tikz}
\usepackage{booktabs,multirow}
\usetikzlibrary{patterns,arrows}
\usetikzlibrary{arrows.meta}
\definecolor{color1}{RGB}{237, 191, 193}
\definecolor{color2}{RGB}{229, 153, 157}
\definecolor{color3}{RGB}{225, 123, 116}
\definecolor{color4}{RGB}{10,10,200}
\definecolor{color5}{RGB}{203,52,38}

\usepackage{pgf}

\usepackage[nocomma]{optidef}

\usepackage{algorithm}
\usepackage{algorithmic}

\usepackage{multicol}
\usepackage{amsthm}

\newcommand{{\ours}}{SFMTL-Graph}

\usepackage{booktabs}
\usepackage{lipsum}
\usepackage{color,soul}

\definecolor{dkgreen}{rgb}{0,0.6,0}
\definecolor{gray}{rgb}{0.5,0.5,0.5}
\definecolor{mauve}{rgb}{0.58,0,0.82}

\begin{document}

\title{Learning to Collaborate Over Graphs: A Selective Federated Multi-Task Learning Approach
}


\author{
	\IEEEauthorblockN{Ahmed Elbakary, Chaouki Ben Issaid, Mehdi Bennis}
    
	\IEEEauthorblockA{Centre for Wireless Communications (CWC), University of Oulu, Finland\\
    Email: \{ahmed.elbakary, chaouki.benissaid, mehdi.bennis\}@oulu.fi}
    
    \thanks{The work is funded by the European Union through the project 6G-INTENSE (G.A no. 101139266).}

}

\maketitle
\begin{abstract}  
We present a novel federated multi-task learning method that leverages cross-client similarity to enable personalized learning for each client. To avoid transmitting the entire model to the parameter server, we propose a communication-efficient scheme that introduces a feature anchor, a compact vector representation that summarizes the features learned from the client's local classes. This feature anchor is shared with the server to account for local clients' distribution. In addition, the clients share the classification heads, a lightweight linear layer, and perform a graph-based regularization to enable collaboration among clients. By modeling collaboration between clients as a dynamic graph and continuously updating and refining this graph, we can account for any drift from the clients. To ensure beneficial knowledge transfer and prevent negative collaboration, we leverage a community detection-based approach that partitions this dynamic graph into homogeneous communities, maximizing the sum of task similarities, represented as the graph edges' weights, within each community. This mechanism restricts collaboration to highly similar clients within their formed communities, ensuring positive interaction and preserving personalization. Extensive experiments on two heterogeneous datasets demonstrate that our method significantly outperforms state-of-the-art baselines. Furthermore, we show that our method exhibits superior computation and communication efficiency and promotes fairness across clients.
\end{abstract}


\input{1-introduction}

\input{2-related_work}

\input{3-system_model}

\input{4-proposed_methods}

\input{5-simulation}
\input{6-conclusion}
\bibliographystyle{IEEEtran}
\bibliography{references} 

\end{document}

%% file: 1-introduction.tex
\section{Introduction}
Federated Learning (FL)~\cite{mcmahan2017communication} has emerged as a significant paradigm in distributed machine learning, leveraging advances in communication networks and machine learning, particularly the success of deep learning methods~\cite{lecun2015deep}. By enabling multiple devices or organizations to train models collaboratively without centralizing the data, FL preserves privacy and reduces the communication costs of gathering all the data in a single data center, making it particularly appealing for applications in sensitive domains such as healthcare~\cite{10288131}, finance~\cite{liu2023efficient}, and applications involving mobile devices~\cite{hard1811federated}, or Industrial Internet of Things (IIoT)~\cite{nguyen2021federated}. While standard FL aims to learn a single global model for all clients, this approach is often suboptimal for scenarios requiring tailored and personalized models. A primary challenge arises from the inherent heterogeneity of client data, leading to significant distribution shifts that traditional FL methods struggle to handle effectively without careful algorithmic design \cite{karimireddy2020scaffold}.  Although FedAvg and similar methods may perform adequately in relatively homogeneous environments, their performance often degrades under non-independent and identically distributed (non-IID) settings~\cite{kairouz2021advances, sattler2019robust}. This limitation becomes particularly pronounced when the models are expected to perform across varied tasks or domains exhibiting substantial differences in underlying data distributions~\cite{deng2020adaptive}. For instance, in a cross-silo FL setting involving different financial institutions, a collaboratively trained fraud detection model~\cite{SachikoKanamori2022, 9730485} might fail to account for significant differences in fraud patterns across regions or countries. These discrepancies can manifest in diverse ways, including seasonal variations or distinct behavioral patterns linked to fraudulent activities. Consequently, a standard FL approach would face difficulties in accommodating such diversity across data silos due to shifts in the underlying data distribution and the specific characteristics of the local data.

Personalized Federated Learning (PFL) offers a framework to address the challenge of data heterogeneity by aiming to learn a personalized model for each client~\cite{li2021ditto, t2020personalized}. PFL approaches typically frame the problem as optimizing a set of local objectives, tailored to each participant client. 
The potential for knowledge transfer across different machine learning tasks, even those with seemingly limited direct overlap, stems from the observation that they often share common underlying structures or computational primitives~\cite{kaiser2017one}. These primitives, although not always readily apparent, can be exploited by techniques like transfer learning or knowledge distillation~\cite{hinton2015distilling}, enabling tasks with limited overlap to benefit from each other. 

Multi-task learning (MTL)~\cite{crawshaw2020multi} is another paradigm explicitly designed to leverage such shared primitives among different tasks. This observation is particularly valuable in heterogeneous FL environments, where data distributions vary significantly across clients, as it provides a mechanism to balance the benefits of global knowledge sharing with the necessity of local model adaptability. Federated Multi-Task Learning (FMTL) extends the concepts of MTL to the FL setting, allowing clients to learn task-specific personalized models while exploiting shared structure and transferring knowledge across related tasks. By modeling each client's learning problem as a distinct task, FMTL facilitates personalized learning tailored to each client's local data distribution while promoting knowledge transfer through collaboration~\cite{sattler2020clustered}. Existing FMTL strategies often achieve this through mechanisms such as shared model components (e.g., feature extractors), explicit modeling of task relationships, and graph-based regularization techniques that structure collaboration based on inferred similarities between clients.

While graph-based regularization has become a popular approach in FMTL~\cite{9975151}, several significant challenges remain. A key challenge is the construction of the task similarity graph itself, which is crucial for quantifying the potential for beneficial collaboration between clients. Existing works often assume predefined or heuristically determined relationships, which may not accurately capture the complex and evolving similarities observed in real-world heterogeneous FL scenarios. Furthermore, even when a similarity graph is used, collaborating with clients that are not sufficiently similar can lead to \emph{negative} knowledge transfer, degrading individual client performance despite the intention of promoting collaboration~\cite{cui2022collaboration}~\cite{wu2023bold}. Ensuring that collaboration is consistently positive and beneficial is difficult but essential for robust personalized performance. To address these challenges, we propose a novel communication-efficient FMTL method that not only dynamically constructs the task similarity graph but also employs a strategic mechanism to ensure positive collaboration. Specifically, we introduce a community detection-based approach to facilitate the formation of communities among highly similar clients, guaranteeing that knowledge transfer occurs primarily within groups where it is most likely to be beneficial, thereby enhancing personalization and robustness. Our method and the steps involved in it are depicted in figure\ref{fig:concept-graph}. The contributions of the paper are as follows
\begin{itemize}
    \item We propose {\ours}, a novel and communication-efficient FMTL algorithm designed for high-performance personalized learning in heterogeneous environments, enabling clients to leverage knowledge from similar peers.
    \item We formulate a community detection-based approach for dynamically forming client communities based on learned task similarity. Collaboration is strategically restricted within these communities to ensure a positive and beneficial knowledge transfer among clients.
    \item We conduct extensive experimental evaluations on two different datasets to analyze the proposed algorithm, quantifying its computation and communication costs. We also investigate key aspects such as fairness and the dynamic nature of the method.
\end{itemize}

%% file: 2-related_work.tex
\section{Related Work}

\begin{figure*}[t]
    \centering
    \begin{subfigure}[b]
    {\textwidth}
        \centering
        \includegraphics[ scale= 0.3]{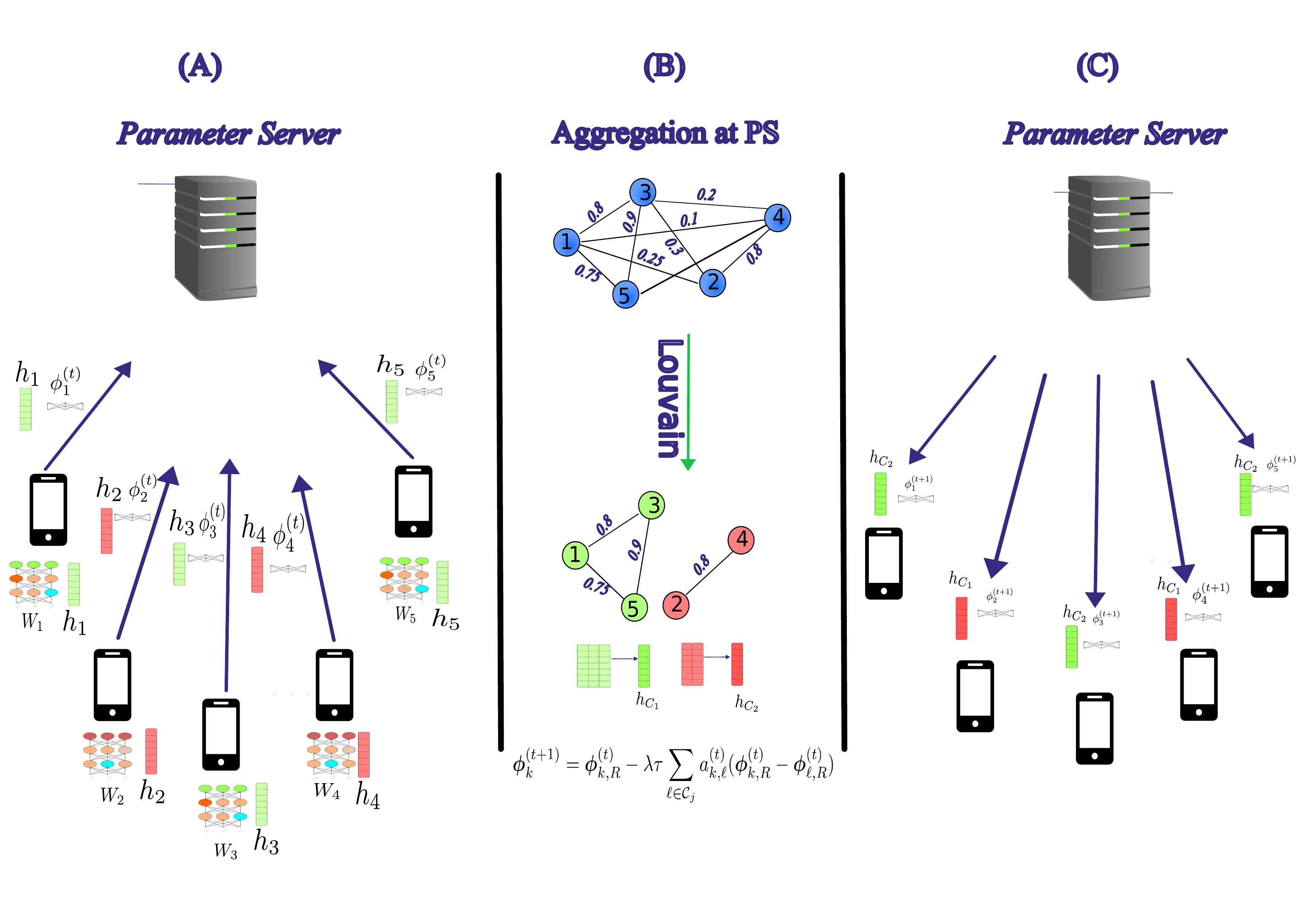}
    \end{subfigure}
    \caption{A schematic view of our proposed method and the details of the steps involved in each stage of the FL standard computation-communication scheme. In the first step (A), clients train a local model with a feature anchor loss and send the classification head $\bm{\phi}$ and the feature anchor $\bm{h}$ to the PS. At the PS, we construct the similarity graph and apply the Louvain method to construct communities of similar clients. After that, we perform an aggregation of both the models, using regularized Laplacian, and the feature anchors using an average of the anchors. All the aggregations are done per community.  The final step is to return those regularized models and aggregated feature anchors to the clients.}
    \vspace{0.5em} 
\label{fig:concept-graph}
\end{figure*}
\subsection{Federated Learning}
Federated Averaging (FedAvg) \cite{mcmahan2017communication} allows different clients to perform local updates for multiple rounds before aggregating the updated local models into a global one. While FedAvg is effective under the assumption that all clients' data distributions are IID, it performs poorly when this assumption is violated. To address the challenges posed by non-IID data distributions, various methods have been proposed. For instance, SCAFFOLD~\cite{karimireddy2020scaffold} incorporates variance reduction techniques to control the divergence of local model updates, allowing FedAvg to achieve more robust performance under distribution shifts. In a similar line of work, Haddadpour et al.~\cite{haddadpour2021federated} proposed using gradient tracking combined with communication compression to simultaneously reduce communication costs and the adverse effects of data heterogeneity. However, a fundamental limitation inherent to methods that train a single global model for all clients lies in their inherent inability to fully personalize the learned model to meet the specific requirements and local data characteristics of individual participant clients. 
\subsection{Personalized Federated Learning}
PFL is a rapidly evolving area specifically aimed at addressing the limitations of learning a single global model by enabling each client to learn a personalized model. PFL encompasses a wide range of algorithms and design choices. One promising class of techniques addresses the PFL problem by decoupling the client model into a shared component, typically a feature extractor, and a personalized component, typically a classifier head. These methods aim to align the shared components to learn a common, useful representation space while allowing clients to personalize their classification heads for their specific local tasks. Building upon this concept, Xu et al.~\cite{xu2023personalized} propose computing a static feature centroid locally for each class in a client's dataset, summarizing learned features. This centroid is transmitted to the server along with the updated client's model. The server aggregates these centroids to form a global centroid, which is then broadcast back to clients, allowing them to align their local feature extractors towards this global representation while maintaining personalized classifiers. Extending this direction, Zhou et al.~\cite{zhou2023fedfa} employ a learned feature anchor instead of a static centroid and perform feature alignment between the global (averaged) feature anchor and the local feature extractor. It also leverages classifier calibration to prevent deviations across clients. While these approaches significantly advance model personalization, a notable limitation is the substantial communication overhead they often incur, typically requiring clients to share updates for the full local model parameters in addition to the feature summary vector (centroid or anchor).
\subsection{Federated Multi-Task Learning}
FMTL offers another powerful perspective for addressing PFL challenges by explicitly modeling each client's learning problem as a distinct task. This framework provides flexibility to adapt concepts from centralized MTL to the FL setting. Early work in FMTL, such as that by Smith et al.~\cite{smith2017federated}, formulated the problem using regularization functions defined by a similarity matrix to capture relationships between clients' tasks. The authors in \cite{marfoqFederatedMultiTaskLearning} proposed an Expectation-Maximization (EM)-based algorithm for FMTL, treating each client's data distribution as a mixture. However, similar to many PFL algorithms discussed earlier, these early FMTL methods often incur significant communication costs, hindering their reliability in resource-constrained environments. Another common approach for FMTL employs Laplacian regularization, aligning locally learned models towards a common space based on clients' similarities~\cite{9975151}.  More recently, the authors in \cite{issaid2025tackling} explored sheaf-based methods, extending graph Laplacian concepts to model relationships between client models via a sheaf over the graph, a technique capable of handling heterogeneity in model architectures. However, this sheaf-based approach appears better suited for cross-silo settings with more computational resources than typical resource-constrained edge devices. While promising, graph-based FMTL methods face two significant challenges. First, determining the appropriate similarity matrix or graph structure dynamically to accurately reflect the relationships between clients. Existing work often relies on predefined relationships or simple and static heuristic approaches, which may fail to capture the dynamic and complex nature of real-world client similarities. Second, ensuring that the graph structure facilitates \emph{positive} knowledge transfer and actively prevents performance degradation caused by collaboration with dissimilar tasks. We address these challenges by dynamically adapting the similarity graph over communication rounds using a similarity-aware metric and by introducing a community detection-based approach to ensure that collaboration occurs only among highly similar clients.

\subsection{Community detection}
Community detection is an algorithmic framework that can identify a set of communities and underlying groups in complex networks~\cite{newman2004detecting}. The overall objective of a community detection method is to find a densely connected subgraph, where each node in the subgraph is highly similar to other members of the community. Community detection algorithms can be beneficial, especially in the case of heterogeneous federated learning. Consider the case where we model the problem using a similarity graph, where each node is a client and each edge is a weight of similarity between those clients. Such a system is highly non-homogeneous, which means that the similarity graph does not form a single coherent group, but a set of distinct ones. In the world of FL, community detection has been a key tool in various works, often as a tool for identifying a set of relevant clients and learning a single model per such a community~\cite{bettinelli2024discovering}. We extend this idea to maintain a coherent and positive collaboration among the detected communities.

In summary, there is a need for a communication-efficient FMTL approach that can dynamically capture client similarities and strategically manage collaboration to ensure positive knowledge transfer, thereby addressing the limitations of existing global FL, PFL, and FMTL methods. Our proposed method builds upon the strengths of FMTL while tackling these specific challenges using a novel combination of communication-efficient client representations, dynamic graph construction, and a community detection-based mechanism.

%% file: 3-system_model.tex
\section{System Model and Problem Formulation}\label{systemModel}
We consider a network of $K$ clients, where the set of all clients is defined as $\mathcal{N} = \{1, \dots, K\}$. Each client communicates only with the PS, which is responsible for coordinating the collaborative learning process. Each client $k$ is associated with a distinct task $t_k$ and holds a local dataset $(\bm{X}_k, \bm{y}_k)$ of $n_k$ samples, where $\bm{X}_k$ represents the input data and $\bm{y}_k$ the corresponding labels. The client's local objective is to train a model $\bm{w}_k \in \mathbb{R}^{d}$ by minimizing a local loss function $f_k({\bm{w}_k}): \mathbb{R}^d \to \mathbb{R}$ for $R$ local rounds. Each client $k$ has a total of $n_k$ samples. At each communication round $t$, the PS randomly samples a subset of clients $\mathcal{S}^{(t)} \subseteq \mathcal{N}$ to participate in the training. Let $m^{(t)} = \sum_{k \in \mathcal{S}^{(t)}} n_k$ be the total number of training samples across all participant clients in round $t$. 

FMTL aims to train these local models to be highly personalized while simultaneously enabling beneficial knowledge transfer across related tasks. A common approach for FMTL frames the problem as minimizing a global objective function that balances local task performance with regularization terms encouraging collaboration
\begin{align}\label{regopt}
\min_{\bm{W}} J(\bm{W}) = F(\bm{W)} + \lambda \mathcal{R}(\bm{W}),
\end{align}
where $\bm{W} = [\bm{w}_{1}^{\top}, \dots, \bm{w}_{K}^{\top}] \in \mathbb{R}^{dK}$ denotes the concatenated vector of model parameters from all $K$ clients and $\lambda \ge 0$ is a regularization parameter controlling how much collaboration is encouraged. The function $F$ is defined as
\begin{align}
    F(\bm{W}) = \frac{1}{K} \sum_{k=1}^{K} f_k(\bm{w_k}),
\end{align}
and the function $\mathcal{R}$ is the regularization function that encourages clients to collaborate by regularizing the models' weights into a closer space. A larger $\lambda$ implies stronger regularization and encourages clients' models to be more similar. When $\lambda = 0$, this objective simplifies to independent local training, which does not leverage any shared knowledge.

Inspired by the concept that learning tasks often share an underlying structure \cite{kaiser2017one}, we assume a hidden relationship between clients' tasks and model this relationship using a weighted undirected graph $\mathcal{G}^{(t)} = (\mathcal{N}, \mathcal{E}^{(t)}, \Psi^{(t)})$, where $\mathcal{N}$ is the set of clients, $\mathcal{E}^{(t)} = \{ (k, \ell) \mid k, \ell \in \mathcal{N}, k \neq \ell\}$ represents the set of all possible edges between distinct clients, and $\Psi^{(t)}$ is a function assigning non-negative edge weights $a_{k\ell}^{(t)} \ge 0$ to each edge $(k, \ell)$.  These weights quantify the strength of the relationship or similarity between tasks $k$ and $\ell$. The graph, and thus the adjacency matrix $\bm{A}^{(t)}$ with entries $a_{k\ell}^{(t)}$, is built and dynamically updated at each communication round based on metrics reflecting client data and model similarity. We assume $\bm{A}^{(t)}$ is symmetric, i.e., $a_{k\ell}^{(t)} = a_{\ell k}^{(t)}$.

For the graph $\mathcal{G}^{(t)}$, the set of neighboring clients of a client $k$ is defined as $\mathcal{N}_k = \mathcal{N} \setminus \{k\}$. For this graph $\mathcal{G}$, we define the degree matrix $\bm{D} = \operatorname{diag}[\delta_1, \ldots, \delta_K] \in \mathbb{R}^{K \times K}$, where $\delta_k = \sum_{\ell \in \mathcal{N}, \ell \neq k} a_{k\ell}$ is the weighted degree of client $k$. The unnormalized graph Laplacian is then defined as $\bm{L} = \bm{D} - \bm{A}$. To apply graph regularization across all dimensions of the model parameters, we use the extended Laplacian matrix $\bm{\widetilde{L}} = \bm{L} \otimes \bm{I}_{d}$, where $\otimes$ is the Kronecker product and $\bm{I}_d$ is the $d \times d$ identity matrix.
An established method to employ graph-based regularization in FMTL is to set the regularization function $\mathcal{R}$ as the Laplacian regularization, defined as
\begin{align}\label{reg.term}
    \mathcal{R}(\bm{W})=\bm{W}^T \bm{\widetilde{L}} \bm{W}=\frac{1}{2} \sum_{k=1}^K \sum_{\ell \in \mathcal{N}_k} a_{k \ell}\left\|\bm{w}_k-\bm{w}_{\ell}\right\|^2,
\end{align}
where the norm $\left\| \cdot \right\|$ represents the Euclidean norm of the difference between model parameter vectors.

Each model $\bm{w}_k$ can be conceptually split into two components: a base backbone or feature extractor $\bm{\theta}_k$ and a task-specific head $\bm{\phi}_k$, such that $\bm{w}_k = \{\bm{\theta}_k, \bm{\phi}_k\}$. The classification head $\bm{\phi}_k$ is typically designed to capture the unique information of each client's task, often mapping features learned by the backbone to client-specific output logits. For instance, in convolutional neural networks (CNNs), the classification head is commonly a linear layer applied after the final feature maps. Conversely, the backbone $\bm{\theta}_k$ learns lower-level or intermediate feature representations that are often more similar across clients, particularly those with related tasks. Finally, we define a feature anchor $\bm{h}_k \in \mathbb{R}^{C_k \times d_h}$ for each client $k$, where $C_k$ is the number of local classes at client $k$ and $d_h$ is the dimension of the feature extractor's output. This anchor is a prototype summarizing the learned features for class $c$ at client $k$. We assume that all clients have the same number of local classes $C_k$. This representation consists of a $d_h$-dimensional vector for each local class, summarizing the learned features for that class.

%% file: 4-proposed_methods.tex
\section{Federated Multi-Task Learning over graphs}
\input{algos}

\subsection{Task Similarity Graph Construction}
Based on the general formulation \eqref{regopt}, our approach implicitly drives collaboration through a personalized local training objective and a server-side aggregation mechanism informed by client similarity and community structure. We formulate our problem based on the following regularization function
\begin{align}\label{new-reg-eq}
\mathcal{R}(\bm{\Theta}, \bm{\Phi}) = \sum_{k=1}^{K} \Bigg[ 
    & g_{k}\left({\bm{\theta}_k}\right) + \sum_{\ell \in \mathcal{C}_k} a_{k, \ell} \left\|\bm{\phi}_{k, R}-\bm{\phi}_{\ell, R}\right\|^2  \Bigg],
\end{align}
where $\mathcal{C}_k$ represents the community to which client $k$ belongs (to be defined later in Section \ref{sec:gametheory}), $\bm{\Theta} = [\bm{\theta}_1^{\top}, \dots, \bm{\theta}_k^{\top}]$, $\bm{\Phi} = [\bm{\phi}_1^{\top}, \dots, \bm{\phi}_k^{\top}]$, and $g_{k}$ is the feature anchor loss of client $k$, which encourages the feature extractor $\bm{\theta}_k$ to produce representations close to the current feature anchors $\bm{h}_k^{(t)}$. This alignment is done for each class $c$ in the dataset $\{\bm{h}_{k}^{(c)} \}_{c=1}^{C_k}$. Specifically, $g_k$ is defined as the mean squared error between features extracted from local data and the corresponding class anchors
\begin{align}\label{anchor-loss}
    g_{k}\left(\bm{\theta}_k\right) = \left\|\bm{\theta}_{k, R}(x_i)-\bm{h}_{k, R}^{(c)}\right\|^2,
\end{align}
where $\bm{h}_{k, R}^{(c)} \in \mathbb{R}^{d_h}$ is the feature anchor of client $k$ for local class $c$. This regularization ensures that we apply it to both the classification head and the feature extractor. We hypothesize that the representation that comes out of the feature extractor $\bm{\theta}_k$ captures the client's local data distribution better. Since sending this feature extractor to the server is not an ideal solution due to the communication cost, we need to find a balance between communication efficiency and performance. Instead of sending the whole model to the PS, we send a feature anchor that can capture these low-level details of the backbone feature extractor. The regularized update contains two parts, one that requires information exchange between clients and one that can be done locally. Hence, the regularization is carried out in two steps, the first at the client locally and the second at the server. To this end, at communication round $t$, we solve the following problem for each selected client
\begin{align}
    \min_{\bm{w}_k}\mathcal{L}_k\left(\bm{w}_k\right)\triangleq \mathbb{E}_{(\bm{x}, \bm{y}) \in \mathcal{D}_k}\left[f_k(\bm{w}_k)+\lambda g_{k}\left(\bm{\theta}_k\right)\right].
\end{align}
This local optimization problem is solved by each selected client $k \in \mathcal{S}^{(t)}$ using $R$ steps of Mini-batch Gradient Descent (GD) or its variants. After $R$ local steps, client $k$ obtains updated model parameters $\bm{w}_{k, R}^{(t)}$ consisting of the feature extractor $\bm{\theta}_{k, R}^{(t)}$ and the classification head $\bm{\phi}_{k, R}^{(t)}$. To facilitate communication efficiency and server-side operations, client $k$ then sends its updated classification head $\bm{\phi}_{k, R}^{(t)}$ and its newly computed feature anchor $\bm{h}_{k, R}^{(t)}$ to the server.

Upon receiving the classification heads and feature anchors from all sampled clients in $\mathcal{S}^{(t)}$, the server constructs a weighted undirected similarity graph among these clients. For each pair of clients $(k, \ell) \in \mathcal{S}^{(t)}$, the similarity weight $a_{k \ell}^{(t)}$ is computed as a combination of their classification head similarity and feature representation similarity 
\begin{align}\label{akl-eq}
    a_{k \ell} ^{(t)}= \alpha \cdot \operatorname{Sim}_{\text {head }}(k, \ell)+(1-\alpha) \cdot \operatorname{Sim}_{\mathrm{repr}}(k, \ell),
\end{align} 
where $\alpha \in [0, 1]$ is a trade-off hyperparameter that balances how much importance to give to the representation similarity over the classification head similarity. The two terms measure similarity in both the classification head and the representations and are defined as
\begin{align}
    \operatorname{Sim}_{\text {head }}(k, \ell) = \frac{1}{2} \sum_{\bm{h} \in \{\bm{h}_{k, R}^{(c)}, \bm{h}_{\ell, R}^{(c)}\}} \operatorname{SimCos} (\bm{\phi}_k(\bm{h}), \bm{\phi}_\ell (\bm{h})).
\end{align}
Note that this expression is done for every class $c$ of client $k$ and client $\ell$. This measure quantifies how similarly the classification heads of clients $k$ and $\ell$ respond to the feature anchors from \emph{both} clients $\ell$ and $k$ for shared classes $c$. In other words, it compares the logits produced by the two classification heads under the same input vector. This approach captures functional similarity. Initially, we considered using the negative $L_1$ norm between the classification head parameters ($\bm{\phi}_k, \bm{\phi}_\ell$) as a similarity measure. However, this approach quantifies parameter similarity, which does not always reflect functional similarity, i.e., how the heads actually behave. For instance, two clients' classification heads might have significantly different parameter values but produce very similar outputs (logits) when applied to the same input features. Since our feature anchors ($\bm{h}_k$) are compact representations of learned features, a more relevant measure of head similarity is how similarly they process these key representations. This functional comparison, specifically comparing their outputs on inputs derived from the feature space, is better captured by the cosine similarity operator, $\operatorname{SimCos}(\cdot,\cdot)$, defined as
\begin{align}
    \operatorname{SimCos}(\bm{z}_k,\bm{z}_\ell) = \frac{\bm{z}_k \cdot \bm{z}_{\ell}}{\|\bm{z}_k\| \|\bm{z}_{\ell}\|},
\end{align}
where $\bm{z}_k$ and $\bm{z}_\ell$ are two arbitrary vectors of the same shape. To measure the similarity in the representation space between the two clients, we compare the two feature anchors of each local class directly using the $\operatorname{SimCos}$ operator as follows
\begin{align}
    \operatorname{Sim}_{\mathrm{repr}}(k, \ell) = \operatorname{SimCos} (\bm{h}_{k, R}^{(c)},  \bm{h}_{\ell, R}^{(c)}).
\end{align}
To prevent negative weights between two clients, we take the maximum of the resultant weight and $0$, i.e., $a_{k, \ell}^{(t)} = \max (a_{k, \ell}^{(t)}, 0)$.
Once the dynamic similarity graph $\mathcal{G}^{(t)}$ is constructed, we formulate the problem as a community detection problem to determine beneficial collaboration structures.
{\subsection{Community Formation}\label{sec:gametheory}
Given the task similarity graph, we need to find the most similar communities, where collaboration is beneficial to the clients. In other words, we aim to generate a set of disjoint groups/communities of clients such that the similarity is maximized. This situation is almost equivalent to a clique partitioning problem~\cite{miyauchiExactClusteringInteger2018}. A clique of a graph is a subset of vertices in a graph where every vertex in the subset is connected to every other vertex in the subset. A clique partitioning problem tries to find the smallest number of cliques in a graph such that every vertex in the graph is represented in exactly one clique. ~\cite{bhasker1991clique}. The resultant partitioning of the graph is a set of disjoint groups, each with its own set of nodes/clients, where each client belongs to one group/community only.  

Since finding the optimal partitioning of the graph $\mathcal{G}^{(t)}$ is known to be an NP-hard problem, we aim to find a partition that maximizes the modularity score $Q$, which evaluates the quality of a graph partitioning by comparing the density of connections within communities to the density of connections between them~\cite{yang2013networks}. For our weighted undirected graph $\mathcal{G}^{(t)}$ with total weight $M^{(t)} = \frac{1}{2} \sum_{k, \ell \in \mathcal{S}^{(t)}, k \neq \ell} a_{k\ell}^{(t)}$, the modularity $Q^{(t)}$ is defined as
\begin{align}
    Q^{(t)}=\frac{1}{2 M^{(t)}} \sum_{(k, \ell) \in \mathcal{C}_k^{(t)}}\left[a_{k\ell}^{(t)}-\frac{\delta_k^{(t)} \delta_\ell^{(t)}}{2 M^{(t)}}\right] \mathbb{I}\left(\mathcal{C}_k^{(t)}, \mathcal{C}_{\ell}^{(t)}\right),
\end{align}
where $\mathbb{I}\left(\mathcal{C}_k^{(t)}, \mathcal{C}_{\ell}^{(t)}\right)$ is an indicator function that is $1$ if $k$ and $\ell$ belongs to the same community and $0$ otherwise. Maximizing $Q^{(t)}$ identifies partitions where intra-community similarity is high compared to what would be expected by setting the similarity weights randomly.

Since maximizing modularity exactly is computationally intractable for large graphs, we employ the Louvain algorithm~\cite{blondelFastUnfoldingCommunities2008}, a widely used heuristic for community detection that efficiently approximates the optimal partitioning for the problem
\begin{align}\label{game-eq}
    \mathcal{C}^{(t)}_{\star}=\arg \max _{\mathcal{C}^{(t)}} Q^{(t)},
\end{align}
where $\mathcal{C}^{(t)}_{\star}$ is the optimal partitioning of the clients at round $t$. In other words, we seek the partitioning of the graph that maximizes the modularity. The Louvain algorithm is an iterative process that proceeds in phases. Initially, each client in $\mathcal{S}^{(t)}$ is placed in its own singleton community. In the first phase, the algorithm iteratively considers each client and evaluates the modularity gain $\Delta Q$ that would result from moving it to a neighboring client's community. A client joins the community that yields the largest positive $\Delta Q^{(t)}$. This process is repeated until no moves can improve the total modularity. The modularity gain $\Delta Q^{(t)}$ for moving client $k$ into a community $\mathcal{C}_j^{(t)}$ is given by
\begin{align}\label{eq-mod-gain}
    \Delta Q^{(t)}=\left[\frac{\sum_{in}^{(t)}+\delta_{k, in}^{(t)}}{2 M^{(t)}}-\left(\frac{\sum_{tot}^{(t)}+\delta_k^{(t)}}{2 M^{(t)}}\right)^2\right] \nonumber \\-\left[\frac{\sum_{in}^{(t)}}{2 M^{(t)}}-\left(\frac{\sum_{tot}^{(t)}}{2 M^{(t)}}\right)^2-\left(\frac{\delta_k^{(t)}}{2 M^{(t)}}\right)^2\right],
\end{align}
where $\sum_{in}^{(t)}$ is the sum of the weights of all the clients inside community $\mathcal{C}_j^{(t)}$, $\sum_{tot}^{(t)}$ is the sum of the weights of all the neighbors of clients in the community $\mathcal{C}_j^{(t)}$, and $\delta_{k, in}^{(t)}$ is the sum of the weights of all neighbors of $k$ inside the community $\mathcal{C}_j^{(t)}$. We assign client $k$ to the community $\mathcal{C}_j^{(t)}$ for which \eqref{eq-mod-gain} is maximum. In the second phase, the identified communities are treated as new nodes in a condensed graph, and the first phase is reapplied. This process of optimization and graph refinement is repeated until no further increase in modularity is possible. Algorithm \ref{louvian-algo} summarizes the Louvain method. The output of this step is the partition $\mathcal{C}^{(t)} = \{\mathcal{C}_1^{(t)}, \mathcal{C}_2^{(t)}, \dots\}$ of $\mathcal{S}^{(t)}$. One big advantage of the Louvain algorithm is that its time complexity is generally considered $O(n \log n)$, where $n$ is the number of nodes of the graph, making it a practical and efficient way of finding a good partitioning of a graph.

Once the communities are formed for the current round, the server facilitates collaboration \emph{within} these communities. To quantify the importance and contribution of each client within its community, we use a simple and computationally efficient aggregation mechanism based on the formed communities. We aggregate the feature anchors weighted based on the number of local samples. For the classification heads, we perform a regularized update based on the update defined in \eqref{new-reg-eq}, for all clients in the same community. We end up with one representation per community, where each client contribution or payoff is weighted by the number of data points such that
\begin{align}\label{eq:agg-repr}
    \bm{h}_{\mathcal{C}_{j}}^{(t)} = \sum_{k \in \mathcal{C}_j} \frac{n_k}{m_{\mathcal{C}_j}^{(t)}} \bm{h}_k^{(t)},
\end{align}
where $m_{\mathcal{C}_j}$ is the total number of samples for all clients in community $\mathcal{C}_j$. For the classification heads, the server performs a regularized update for each client $k \in \mathcal{S}^{(t)}$ as follows
\begin{align}\label{reg-update-eq}
    \bm{\phi}_k^{(t+1)} = \bm{\phi}_{k, R}^{(t)} - \lambda \tau \sum_{\ell \in \mathcal{C}_{j}} a_{k, \ell}^{(t)} (\bm{\phi}_{k,R}^{(t)} - \bm{\phi}_{\ell, R}^{(t)}),
\end{align}
where $\tau$ is a server-side learning rate defined as $\tau = \eta \times R$. This update effectively pulls the head of client $k \in \mathcal{C}_j^{(t)}$ towards the heads of other clients $\ell$ \emph{in the same community} $\mathcal{C}_j^{(t)}$, weighted by their similarity $a_{k\ell}^{(t)}$.

The server then sends the aggregated feature anchor $\bm{h}_{\mathcal{C}_j}^{(t)}$, for the community $\mathcal{C}_j^{(t)}$ client $k$ belongs to, and client $k$'s updated classification head $\bm{\phi}^{(t+1)}_k$ back to each client $k$. For the next communication round $(t+1)$, each client initializes its local feature anchor as the community anchor received as follows $\bm{h}_k^{(t+1)} = \bm{h}_{\mathcal{C}_j}^{(t)}$, $\forall$ $k \in \mathcal{C}_j$. This implicitly incorporates knowledge from similar clients via the shared representation space. The client also initializes its classification head for the next local training phase with the server-updated head using $\bm{\phi}_{k, 0}^{(t+1)} = \bm{\phi}_{k}^{(t+1)}$. This process is repeated for $T$ communication rounds until convergence.

%% file: algos.tex
\begin{algorithm}[t]
    \caption{SFMTL-Graph}
    \label{alg:fmtl_graph}
    \begin{algorithmic}[1]
        \STATE {\bf Parameters:} learning rate $\eta$, number of local iterations $R$, hyperparameters $(\lambda, \alpha, \eta)$.
        \STATE {\bf Initialize:} $\bm{w_k}^0$, $\bm{h}^{(0)}  \sim \mathcal{N}(0, I) , \mathcal{C}= \{\}$
        \FOR{communication round $t = 0$ to $T$}
        \STATE Server Sample  a subset $\mathcal{S}^{(t)}$ to initiate local training.
        \IF{$t > 0$}
            \STATE send $\bm{h}^{(t)}_{{c}_{j}}$ and $\bm{\phi}_k^{(t)}$ to client $k$; $\forall k \in c_j$ and $\forall j \in c$.
        \ENDIF
            \FOR{\textbf{client} $k \in \mathcal{S}^{(t)}$}
                \IF{$t > 0$}
                    \STATE Initialize classification head $\bm{\phi}_{k, 0}^{(t)} \leftarrow \bm{\phi}_{k}^{(t)}$
                    \STATE Initialize local feature anchor $\bm{h}_{k}^{(t)} \leftarrow \bm{h}^{(t)}_{{c}_{j}}; k \in \mathcal{C}_j$.
                \ENDIF
                
                \FOR{local iterations $r = 0$ to $R$}
                \STATE Compute the Supervised loss $f_k(\bm{w}_{k, r}^{(t)})$.
                \STATE Compute the anchor loss $g_k(\bm{\theta}^{(t)}_k)$ as in \eqref{anchor-loss}.
                \STATE Update local model using a Mini-batch GD step\\
                
                $\bm{w}_{k, r + 1}^{(t)} \leftarrow \bm{w}_{k, r}^{(t)} - \eta \nabla \mathcal{L}_k(\bm{w}_{k, r}^{(t)})$.
                
                \ENDFOR
                \STATE Send $\bm{\phi}^{(t)}_{k, R}$ and $\bm{h}_{k, R}^{(t)}$ to the server.  
            \ENDFOR
            
            \textbf{On Server}\\
            \STATE If Client $k \notin S^{(t)}$, set $\bm{w}_k^{(t+1)} = \bm{w}_{k}^{(t)}$.
            \FOR{Client $(k, \ell) \in \mathcal{S}^{(t)}$}
                \STATE Calculate the similarity weights $a_{k, \ell}$ using \eqref{akl-eq}.
            \ENDFOR
            \STATE Solve the maximization problem \eqref{game-eq} using Algorithm \ref{louvian-algo} to get $\mathcal{C}^{(t)} = \{\mathcal{C}_1^{(t)}, \mathcal{C}_2^{(t)}, \dots\}$. 
            \STATE Update the feature anchor based on \eqref{eq:agg-repr}.
            \STATE Update the classification heads using \eqref{reg-update-eq},  $\forall (k, \ell) \in \mathcal{C}_{j}$.
        \ENDFOR
    \end{algorithmic}
\end{algorithm}

\begin{algorithm}
\caption{Louvain Algorithm}
\label{louvian-algo}
\begin{algorithmic}[1]
\STATE {\bf Parameters:} Weighted graph $\mathcal{G}$ with weights $a_{k,\ell}$

\STATE {\bf Initialize:} each node in its own community

\REPEAT
    \STATE improvement $\leftarrow$ \textbf{true}
    \WHILE{improvement}
        \STATE improvement $\leftarrow$ \textbf{false}
        \FORALL{$k \in \mathcal{G}$ (in random order)}
            \STATE $c_k \leftarrow$ current community of $k$
            \STATE $\Delta Q_{\max} \leftarrow 0$, $c_k^\ast \leftarrow c_k$
            \FORALL{$\ell \in \mathcal{N}_k$}
                \STATE $c_\ell \leftarrow$ community of $\ell$
                \STATE Compute $\Delta Q$ (Eq.\eqref{eq-mod-gain}) if $k$ moved to $c_\ell$
                \IF{$\Delta Q > \Delta Q_{\max}$}
                    \STATE $\Delta Q_{\max} \leftarrow \Delta Q$
                    \STATE $c_k^\ast \leftarrow c_\ell$
                \ENDIF
            \ENDFOR
            \IF{$c_k^\ast \neq c_k$}
                \STATE Move $k$ to $c_k^\ast$
                \STATE improvement $\leftarrow$ \textbf{true}
            \ENDIF
        \ENDFOR
    \ENDWHILE

    \STATE Coarsen $\mathcal{G}$: 
    \STATE \hspace{1em} Nodes $\rightarrow$ communities, 
    \STATE \hspace{1em} Edge $a_{k,\ell} \rightarrow$ sum over inter-community weights
    \STATE Update $\mathcal{G}$ accordingly

\UNTIL{$\Delta Q$ negligible}

\RETURN Final community labels
\end{algorithmic}
\end{algorithm}

%% file: 5-simulation.tex
\begin{figure*}[t]
    \centering
    \begin{subfigure}[b]
    {0.30\textwidth}
        \centering
        \includegraphics[ scale= 0.65]{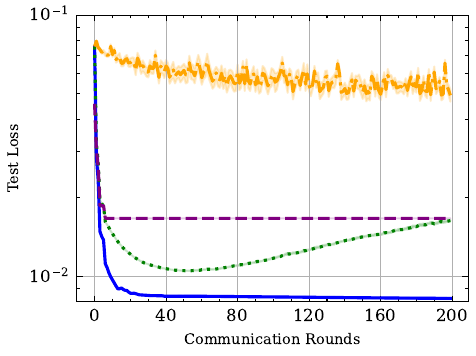}
        \caption{ }
        \label{subfig:test-loss-cifar}
    \end{subfigure}
    \begin{subfigure}[b]
    {0.30\textwidth}
        \centering
        \includegraphics[scale= 0.65]{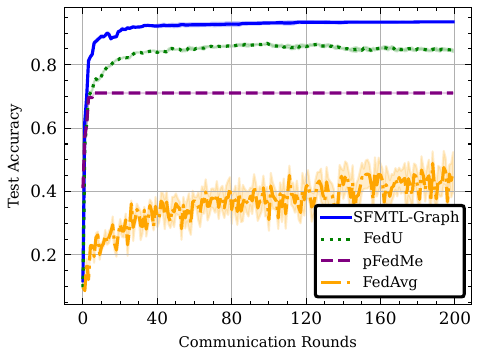}
        \caption{ }
        \label{subfig:test-acc-cifar}
    \end{subfigure}
    \begin{subfigure}[b]
    {0.30\textwidth}
        \centering
        \includegraphics[scale= 0.65]{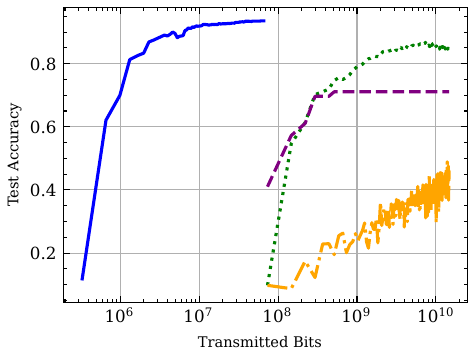}
        \caption{ }
        \label{subfig:comm-cifar}
    \end{subfigure}
    \caption{Heterogeneous CIFAR-10 results}
    \vspace{0.5em} 

    \begin{subfigure}[b]
    {0.30\textwidth}
        \centering
        \includegraphics[scale=0.65]{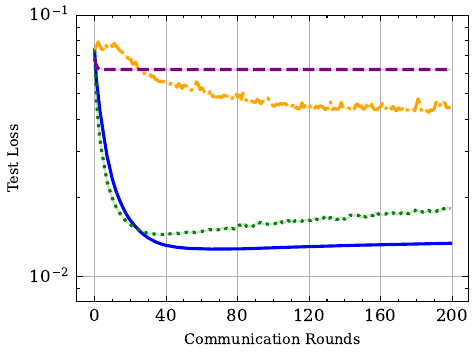}
        \caption{ }
        \label{subfig:test-loss-mnist}
    \end{subfigure}
    \begin{subfigure}[b]
    {0.30\textwidth}
        \centering
        \includegraphics[scale=0.65]{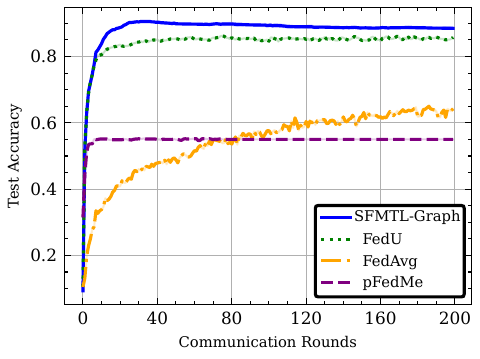}
        \caption{ }
        \label{subfig:test-acc-mnist}
    \end{subfigure}
    \begin{subfigure}[b]{0.30 \textwidth}
        \centering
        \includegraphics[scale= 0.65]{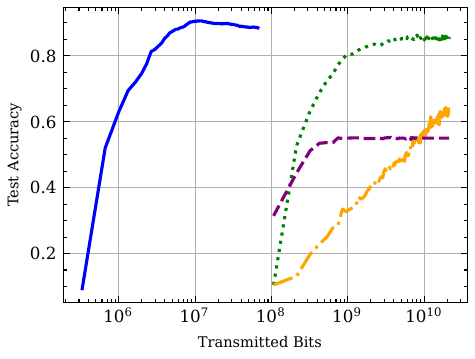}
        \caption{ }
        \label{subfig:comm-mnist}
    \end{subfigure}
    \caption{Rotated \& Masked MNIST results}
    \label{fig:comparison}
\end{figure*}

\section{Experiments}\label{RES}
This section presents the experimental evaluation of {\ours}, detailing the setup, datasets, comparison with baseline methods, and analysis of its key characteristics including performance, communication and computation efficiency, dynamic behavior, and fairness.

\subsection{Experimental Settings}
We evaluate the performance and effectiveness of our proposed {\ours} method using two distinct image classification datasets, designed to represent challenging heterogeneous FL scenarios:
\begin{itemize}
    \item \textbf{Heterogeneous CIFAR-10 Dataset}. A federated learning version of CIFAR-10\cite{krizhevsky2009learning} where each client has a total of two classes to account for non-IIDness of the data, especially an extreme label distribution shift across clients. We consider $30$ clients in total, and we assume the same model is used across all clients. Our model is a simple CNN with three convolutional layers and two fully connected layers. 
    \item \textbf{Rotated \& Masked MNIST}. We consider $40$ clients for this dataset. Based on the original MNIST dataset~\cite{lecun1998gradient}, this dataset adds a feature distribution shift, also known as covariate shift, across clients by applying a rotation on the client's data of $\{0^\circ, 90^\circ, 180^\circ, 270^\circ\}$ degrees and then applying a $2 \times 2$ mask to the resultant rotated data. The model trained by the clients for this dataset is a CNN with two convolutional layers.
\end{itemize}
All algorithms are implemented using PyTorch (version 2.5) and evaluated on a single Nvidia A100 GPU. The primary evaluation metric is average test accuracy across clients. Results are reported as the mean and standard deviation over five independent runs. The code implementation is publicly available at \url{https://github.com/Ahmed-Khaled-Saleh/fmtl-graph/tree/main}. The datasets used are preprocessed and can be automatically downloaded when reproducing the results. 

We compare our method against three different algorithms that span different categories: FedAvg~\cite{mcmahan2017communication} (typical FL), pFedMe~\cite{t2020personalized} (PFL), and FedU~\cite{9975151} (FMTL). For local training, all clients perform $R=5$ local rounds using mini-batch GD with a batch size of $32$. An exception is pFedMe on Rotated \& Masked MNIST, which uses $20$ local rounds. Note that, to perform a fair comparison, we make $30$ inner rounds for pFedMe to make the algorithm converge. Learning rates were tuned for each algorithm to achieve optimal performance: $0.05$ for {\ours} and FedU, $0.005$ for pFedMe, and $0.03$ for FedAvg. The trade-off hyperparameter $\alpha$ in {\ours} is set to $0.49$. We conducted hyperparameter tuning for all methods to ensure a fair comparison using their best respective configurations. 


\subsection{Performance Comparison}
Figures \ref{subfig:test-acc-cifar} and \ref{subfig:test-loss-cifar} present the average test accuracy and loss curves for the Heterogeneous CIFAR-10 dataset. Our method demonstrates significantly superior performance, achieving substantially higher accuracy than all baselines. The closest baseline to our method, FedU, is around $9\%$ less in accuracy. pFedMe's performance is comparable to FedU's, illustrating the strong capabilities of personalized and multi-task learning approaches in this highly heterogeneous setting. As expected, FedAvg exhibits the lowest performance, consistent with the known limitations of traditional FL methods on non-IID data. 

For Rotated \& Masked MNIST, the learning curves are shown in Figures \ref{subfig:test-loss-mnist} and \ref{subfig:test-acc-mnist}. Our algorithm achieves the best performance, although the gap between the closest baseline (FedU) and our method is relatively lower than the CIFAR-10 case. This reduced gap may be attributed to the less extreme heterogeneity of the MNIST-based task compared to the CIFAR-10 scenario. Nevertheless, our algorithm is the best-performing method across both datasets.

\subsection{Computation and Communication Costs}

        

This section quantifies the communication and computation costs of {\ours} and compares them against the baselines.  We illustrate the trade-off between communication cost and performance by tracking the cumulative number of transmitted bits across communication rounds. We assume model parameters are represented using a 32-bit floating-point format. The feature anchor $\bm{h}_k$ has dimensions $C_k \times d_h$, where $d_h=512$. This means that for our method, sharing the feature anchor with the server where the client has two classes locally is equivalent to sending a lightweight vector of size $512 \times 2 = 1024$. Additionally, clients transmit the classification head $\bm{\phi}_k$ of size $512 \times 10$. For the Heterogeneous CIFAR-10 dataset, {\ours} demonstrates significantly lower communication costs than all baselines as illustrated in  Figure \ref{subfig:comm-cifar}. Over $200$ communication rounds, our method transmits less than $10^8$ bits cumulatively. In contrast, FedU, FedAvg, and pFedMe transmit over $10^{10}$ bits, while achieving lower accuracy. The total cumulative bits transmitted by {\ours} over $200$ rounds is even less than the transmission required by any baseline in a single communication round. This highlights the substantial communication efficiency of our method. For the Rotated \& Masked MNIST, as illustrated in Figure \ref{subfig:comm-mnist} the communication cost of our method is still less than any other baseline. Our method transmits almost the same number of bits transmitted in the Heterogeneous CIFAR-10, $10^{8}$ bits. 


Regarding the computation cost, we measure the floating-point operations (FLOPS) of all algorithms based on the amount of local computations needed per round. Our method exhibits the same number of FLOPS as FedU and FedAvg, with around $1.22 \times 10^8$ for Rotated and Masked MNIST and $1.25 \times 10^8$ for the Heterogeneous CIFAR-10. In contrast, pFedMe's local optimization, including its inner loops, required significantly more computation, has around $3.68\times 10^9$ and $1.5\times 10^{10}$ FLOPS for the same two datasets, respectively. This shows clearly that for the same number of local computations and under the same computational complexity, we can outperform all other baselines.

\subsection{Dynamicity of the Task Similarity Graph}

\begin{figure*}
    \centering
    \includegraphics[width=\linewidth, scale=0.45]{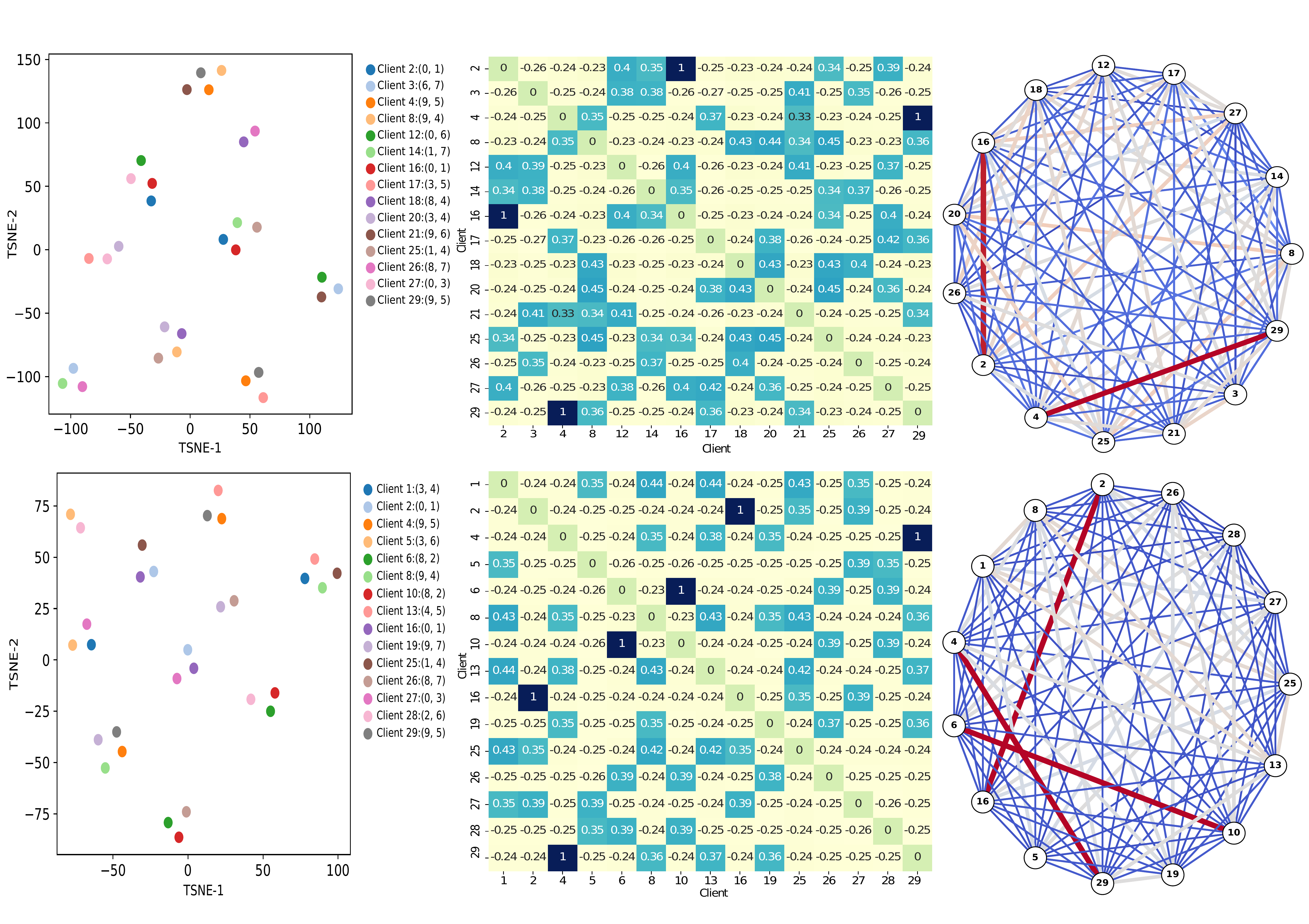}
    \caption{The resultant representation alignment , projected into 2D space using t-SNE, and the classification head response simmilairty matrix, along with the dynamic task similairty graph for communication rounds $t = 50$ (upper) and $t = 150$ (lower). As we can see, there is a strong correlation between the reprentation alignment and the classification head response similairty. A stronger redness in the dynamic similiarty graph (lower and upper right) indicates a stronger similairty.
}
    \label{fig:task-sim}
\end{figure*}

\begin{table}
    \centering
    \caption{communities formed across four different rounds in the Heterogeneous CIFAR-10 dataset.}
    \label{table-coalitions}
    \begin{tabular}{ccc}
        \toprule
        \textbf{Number of communities} & \textbf{Round} & \textbf{Communities} \\
        \midrule
          & & $\{8, 17, 20, 4, 21, 29\}$, \\
          $3$& $50$ & $\{14, 18, 26, 25, 3\}$,\\
          & & $\{27, 12, 16, 2\}$\\
          \midrule
          $2$& $100$& $\{9, 17, 24, 11, 8, 13, 19, 29, 21\}$, \\
          & & $\{7, 23, 10, 27, 22, 2\}$,\\
          \midrule
          & & $\{25, 27, 2, 1, 16, 5\}$, \\
          $3$& $150$ & $\{28, 26, 6, 10\}$,\\
          & & $\{8, 4, 29, 19, 13\}$\\
         \midrule
         & & $\{18, 8, 19, 4, 10, 6, 1\}$,\\
         $3$& $200$ & $\{15, 5, 12, 3\}$,\\
         & & $\{16, 14, 24, 7\}$\\
          \bottomrule
    \end{tabular}
\end{table}
This section empirically investigates the dynamic nature of the task similarity graph and community formation in {\ours}. Since we defined a client or a task as a set of a model and data, it is a natural approach to construct the task similarity graph based on both clients' models and data. As detailed in Section \ref{systemModel}, the graph is constructed based on similarities between clients' feature anchors ($\bm{h}_k$) and classification heads ($\bm{\phi}_k$), dynamically updated each round for sampled clients. We need either a similarity metric or a distance metric that can tell us how close or far those learned representations and the models are. For the feature anchors, we used cosine similarity between vectors to quantify the similarity. We examine the evolution of these similarities and the resulting graph structure every $50$ communication rounds. We project the learned feature representations, or feature anchors, of all clients into a 2D space using t-SNE~\cite{van2008visualizing} to see exactly how features are aligned across clients. In the communication round $50$, we can see that clients that share the same task, for example, client $2$ and client $16$, whose labels are the same, end up very close to each other in the representation space. This means that the underlying model is learning almost the same representations. Even more, the classification head similarity is $1$, which means that the two models are responding equivalently to the same input vector. The represntation and model rsponse similairty for $t= 50$ and $t = 150$ are depicted at the upper and lower part (left and middle) of figure \ref{fig:task-sim}.

To examine another case where two clients only share one class, we consider clients $4$ and $8$. We can observe that the learned representations for the two clients are close only in one case, which is the shared class between them. This aligns with the classification head similarity score, which is only $0.35$. For clients that have one class in common, we can see that most clients' similarity scores for the heads range from $0.30$ to $0.45$. The resultant graphs at rounds $50$ and $150$ are depicted in the upper and lower part (right) of Figure \ref{fig:task-sim}, where a stronger redness of the connection means a stronger relation between the two clients. The communities that are created based on this graph are three communities: $\{8, 17, 20, 4, 21, 29\}$,  $\{14, 18, 26, 25, 3\}$, and $\{27, 12, 16, 2\}$. We can see that the clients who have the same classes end up in the same community as expected. All communities in the four observed rounds are depicted in Table \ref{table-coalitions}. In each iteration, this task similarity graph is updated to reflect the client's local updates. One thing that makes this graph consistent with the actual clients' similarity is that clients who are close to each other, those who share the same set of classes, are pulled closer and closer as the rounds continue. We can examine this by looking at clients $2$ and $16$. If we compare their distance in the representation space from round $50$ to round $150$, we can see that feature anchors are closer in round $150$ than those in round $50$, which indicates that the models are converging to an even closer point. This also indicates that our method tackles heterogeneity of the local data distribution by forming communities of highly similar clients, where the tasks can benefit from each other.
\subsection{Fairness across Clients}
\begin{table}
    \centering
    \caption{Statistics of the test accuracy over $200$ communication rounds on the Heterogeneous CIFAR-10 dataset. We report the mean$\pm$std for each metric.}
    \resizebox{\columnwidth}{!}{%
    \begin{tabular}{ccccc}
        \toprule
        \textbf{Method} &  \textbf{Mean} & \textbf{Worst 10\%} & \textbf{Worst 20\%} & \textbf{STD}\\
        \midrule
          {\ours}& $\bm{0.92}$ $\pm$ $\bm{0.06}$ & $\bm{0.75}$ $\pm$ $ 0.11$ & ${0.78}$ $\pm$ $ 0.10$ &  $\bm{0.08} \pm \bm{0.03}$\\
         Fedu& $0.91$ $\pm$ $ 0.08$ & $0.73$ $\pm$ $ 0.13$ & $\bm{0.78}$ $\pm$ $ \bm{0.12}$ &  $0.09 \pm 0.03$\\
         pFedMe& ${0.78}$ $\pm$ $ 0.03$ & ${0.59}$ $\pm$ $0.07$ & ${0.61}$ $\pm$ $ 0.07$ &  $0.13 \pm 0.02$\\
         FedAvg& ${0.41}$ $\pm$ $ 0.08$ & ${0.08}$ $\pm$ $ 0.08$ & ${0.13}$ $\pm$ $ 0.08$ &  $0.21 \pm 0.03$\\
        \bottomrule
    \end{tabular}%
    }
    \label{tab:fairness}
\end{table}
Fairness is a crucial consideration in FL, assessing how performance is distributed across individual clients, especially in heterogeneous settings where some clients might be disadvantaged. We evaluate fairness by analyzing the distribution of test accuracy across all clients. Specifically, we report the overall mean test accuracy, the average accuracy achieved by the worst-performing 10\% and 20\% of clients, and the standard deviation of test accuracy across all clients over 200 communication rounds on the Heterogeneous CIFAR-10 dataset. This evaluation follows the methodology described in~\cite{liLearningCollaborateDecentralized2022}. These metrics help determine if an algorithm's high average performance is consistent across the client population or skewed, potentially leaving some clients significantly behind. Table \ref{tab:fairness}  presents the fairness statistics. We can clearly see that {\ours} achieves the highest mean accuracy and the lowest standard deviation, , indicating both superior average performance and less variance in performance across the client population compared to baselines. For the worst-performing clients, {\ours} achieves the highest average accuracy for the bottom 10\% of clients. While FedU shows a slightly higher average for the bottom 20\%, the difference is marginal. These results demonstrate that {\ours} promotes fair performance among participant clients, avoiding significant bias or leaving certain clients with substantially lower accuracy, which is a common challenge in heterogeneous FL. 

%% file: 6-conclusion.tex
\section{Conclusion}
In this work, we proposed {\ours}, a novel Federated Multi-Task Learning method designed for personalized learning in heterogeneous environments while ensuring communication efficiency and fostering beneficial collaboration. Our method leverages a dynamic task similarity graph constructed from learned client feature anchors and classification heads. To address the risk of negative knowledge transfer, we introduced a community detection approach utilizing the Louvain algorithm to partition clients into communities based on this graph, restricting collaboration to highly similar peers within these groups. This selective collaboration mechanism, combined with a communication-efficient architecture utilizing feature anchors and lightweight heads, allows {\ours} to effectively balance personalization and knowledge sharing. Extensive experimental evaluation on two heterogeneous datasets demonstrates the significant effectiveness of our method, showing superior performance and communication efficiency compared to baseline approaches across traditional FL, PFL, and FMTL paradigms. Furthermore, our analysis confirms the dynamic nature of the task similarity graph, which adapts to learned representations, and highlights {\ours}'s ability to promote fairer performance distribution across heterogeneous clients.